# Causal discovery of linear acyclic models with arbitrary distributions

**Patrik O. Hoyer**
**Aapo Hyvärinen**

Helsinki Institute for
Information Technology &
Department of Computer Science
University of Helsinki
Finland

**Richard Scheines**
**Peter Spirtes**
**Joseph Ramsey**

Department of Philosophy
Carnegie Mellon University
Pittsburgh, PA, USA

**Gustavo Lacerda**

Machine Learning Department
Carnegie Mellon University
Pittsburgh, PA, USA

**Shohei Shimizu**

Osaka University
Japan

## Abstract

An important task in data analysis is the discovery of causal relationships between observed variables. For continuous-valued data, linear acyclic causal models are commonly used to model the data-generating process, and the inference of such models is a well-studied problem. However, existing methods have significant limitations. Methods based on conditional independencies (Spirtes et al. 1993; Pearl 2000) cannot distinguish between independence-equivalent models, whereas approaches purely based on Independent Component Analysis (Shimizu et al. 2006) are inapplicable to data which is partially Gaussian. In this paper, we generalize and combine the two approaches, to yield a method able to learn the model structure in many cases for which the previous methods provide answers that are either incorrect or are not as informative as possible. We give exact graphical conditions for when two distinct models represent the same family of distributions, and empirically demonstrate the power of our method through thorough simulations.

## 1 INTRODUCTION

In much of science, the primary focus is on the discovery of *causal* relationships between quantities of interest. The randomized controlled experiment is geared specifically to inferring such relationships. Unfortunately, in many studies it is unethical, technically extremely difficult, or simply too expensive to conduct such experiments. In such cases causal discovery must be based on uncontrolled, purely observational data combined with prior information and reasonable assumptions.

In cases in which the observed data is continuous-valued, linear acyclic models (also known as recursive Structural Equation Models) have been widely used in a variety of fields such as econometrics, psychology, sociology, and biology; for some examples, see (Bollen 1989). In much of this work, the structure of the models has been assumed to be known or, at most, only a few different models have been compared. During the past 20 years, however, a number of methods have been developed to learn the model structure in an unsupervised way (Spirtes et al. 1993; Pearl 2000; Geiger and Heckerman 1994; Shimizu et al. 2006). Nevertheless, all approaches so far presented have either required distributional assumptions or have been overly restricted in the amount of structure they can infer from the data. In this contribution we show how to combine the strenghts of existing approaches, yielding a method capable of inferring the model structure in many cases where previous methods give incorrect or uninformative answers.

The paper is structured as follows: Section 2 precisely defines the models under study, and Section 3 discusses existing methods for causal discovery of such models. In Section 4 we formalize the discovery problem and give exact theoretical results on identifiability. Then, in Section 5 we introduce and analyze a method termed `PClingam` that combines the strenghts of existing methods and overcomes some of their weaknesses, and is, in the limit, able to estimate all identifiable aspects of the underlying model. Section 6 provides empirical demonstrations of the power of our method. Finally, Section 7 maps out future work and Section 8 provides a summary of the main points of the paper.

## 2  LINEAR MODELS

In this paper, we assume that the observed data has been generated by the following process:

1. The observed variables $x_i$, $i = \{1 \ldots n\}$ can be arranged in a *causal order*, such that no later variable causes any earlier variable. We denote such a causal order by $k(i)$. That is, the generating process is *recursive* (Bollen 1989), meaning it can be represented graphically by a *directed acyclic graph* (DAG) (Pearl 2000; Spirtes et al. 1993).

2. The value assigned to each variable $x_i$ is a *linear function* of the values already assigned to the earlier variables, plus a 'disturbance' (noise) term $e_i$, and plus an optional constant term $c_i$, that is

$$x_i = \sum_{k(j)<k(i)} b_{ij} x_j + e_i + c_i, \qquad (1)$$

   where we only include non-zero coefficients $b_{ij}$ in the equation.

3. The disturbances $e_i$ are all continuous random variables with arbitrary densities $p_i(e_i)$, and the $e_i$ are independent of each other, i.e. $p(e_1, \ldots, e_n) = \prod_i p_i(e_i)$.

This formulation neither requires the disturbances to be normally distributed nor does it require them to have non-Gaussian (non-normal) densities. In general, some of the distributions can be Gaussian and some not, and we do not a priori know which are which.

We assume that we are able to observe a large number of data vectors $\mathbf{x}$ (which contain the variables $x_i$), and each data vector is generated according to the above described process, with the same causal order $k(i)$, same coefficients $b_{ij}$, same constants $c_i$, and the disturbances $e_i$ sampled independently from the same distributions. Note that the independence of the disturbances implies that there are *no unobserved confounders* (Pearl 2000). Spirtes et al. (1993) call this the *causally sufficient* case.

Finally, we assume that the observed distribution is *faithful* to the generating graph (Spirtes et al. 1993), i.e. the model is *stable* in the terminology of Pearl (2000). If the model parameters are in some sense randomly generated, this is not a strong assumption, as violations of faithfulness have Lebesgue measure 0 in the space of the linear coefficients.

An example of such a model is given in Figure 1a. Note that the full model consists of a directed acyclic graph over the variables, the connection strenghts $b_{ij}$, the constants $c_i$, and the densities $p_i(e_i)$. In this example we have chosen $c_i = 0$ for all $i$, so these are not shown.

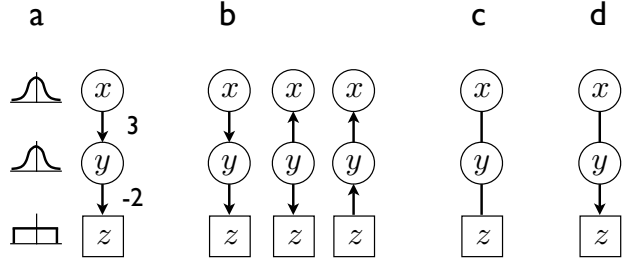

Figure 1: An example case used to illustrate the concepts described in Sections 2–4. **(a)** A linear, acyclic causal model for $x$, $y$ and $z$. The data is generated as $x := e_x$, $y := 3x + e_y$, and $z := -2y + e_z$, with $e_x$ and $e_y$ drawn from Gaussian distributions and $e_z$ from a non-Gaussian distribution, and $e_x$, $e_y$ and $e_z$ are all mutually independent. Note that we show variables with Gaussian disturbances using circles whereas variables with non-Gaussian disturbances are marked by squares. **(b)** The three directed acyclic graphs over $x$, $y$ and $z$ which all entail the same conditional independence relationships as the generating model. **(c)** The three DAGs in (b) succintly represented as a *d-separation-equivalence pattern*. **(d)** The *distribution-equivalence pattern* of the original model.

## 3  EXISTING METHODS

Given our set of data vectors $\mathbf{x}$, to what extent can we estimate the data generating process? Obviously, if the number $N$ of data vectors is small, estimation may be quite unreliable. Therefore we will here mainly focus on the theoretical question: To what extent (and with what methods) can we identify the true model in the limit as $N \to \infty$?

The most well-known approach to inference of this type of causal networks is based on (conditional) independencies between the variables (Spirtes et al. 1993; Pearl 2000). When, as in our case, there are assumed to be no hidden confounding variables and no selection bias, one can in the large-sample limit identify the set of networks which represent the same independencies as the true data generating model. To illustrate, in Figure 1b we show all three DAGs which imply the set of independencies produced by the true model. This set is known as the d-separation-equivalence class, and is often represented in the form of a *d-separation-equivalence pattern*: a partly directed graph in which undirected edges represent edges for which both directions are present in the equivalence class (Spirtes et al. 1993), as illustrated in Figure 1c. We want to emphasize that, using conditional independence information alone, it is impossible to distinguish between members inside a d-separation-equivalence class because these (by definition) represent the same set of conditional independencies between the observed variables.

Fortunately, in many cases there is additional information available that can be used to further distinguish between different DAGs. In particular, it can be shown (Shimizu et al. 2006) that if all (or all but one) distributions of the error variables are non-Gaussian, it is in fact possible to identify the complete causal model, including all the parameters. This is possible using a method based on Independent Component Analysis (ICA) (Hyvärinen et al. 2001). Unfortunately, however, when two or more disturbances are Gaussian the standard method based on ICA will fail. As an extreme example, when all disturbances are Gaussian, standard ICA-based methods return nonsense and are not even able to find the correct d-separation-equivalence class.

These considerations raise the question of whether it is possible to combine the methods so as to obtain robustness with respect to Gaussian distributions but not forgo the possibility of identifying the full model in favourable circumstances. Indeed, such a combination is possible and is presented in Section 5. Here, we simply note that the naïve solution of first running some test and then selecting one of the two methods, will not be optimal. Consider, for instance, our example model in Figure 1a. Because there is more than one Gaussian error variable the standard ICA-based method (Shimizu et al. 2006) is not applicable, and hence one would have to settle for the d-separation-equivalence class (Figure 1c) given by independence-based methods. However, as we show in the next section, in this example we can actually reject one of the DAGs in the equivalence class and hence obtain a smaller set of possible generating models.

## 4 DISTRIBUTION-EQUIVALENCE

First, we need to extend a DAG object to include information on the non-Gaussianity of associated disturbance variables.

**Definition 1** *An **ngDAG** is a pair $(G, ng)$ where $G$ is a directed acyclic graph over a set of variables $V$ and $ng$ is a binary vector of length $|V|$, each element of which is associated with one of the variables of $V$.*

**Definition 2** *We say that a linear acyclic causal model $M$ instantiates an **ngDAG** $D$ (alternatively, $D$ represents $M$) if and only if the directed acyclic graph associated with $M$ is equal to that specified in $D$, and further if the set of variables with non-Gaussian disturbance variables in $M$ is equal to the set of positive entries in the binary vector specified in $D$.*

In general, an ngDAG $D$ is instantiated by many different models $M$ which differ in their connection strengths $b_{ij}$ as well as in their distributions $p_i(e_i)$. Next, we define the important concept of distribution-equivalence between ngDAGs, which defines to what extent it is possible to infer the ngDAG which represents the true data generating causal model, from observational data alone.

**Definition 3** *Two **ngDAGs** $D_1$ and $D_2$ are distribution-equivalent if and only if for any linear acyclic causal model $M_1$ which instantiates $D_1$ there exists an instantiation $M_2$ of $D_2$ which yields the same joint observed distribution as $M_1$, and vice versa.*

Distribution-equivalence partitions the set of ngDAGs into distribution-equivalence classes, and these may be represented using simplified graphs:

**Definition 4** *An **ngDAG** pattern representing an **ngDAG** $D$ is a mixed graph (consisting of potentially both directed und undirected edges), obtained in the following way:*

1. *Derive the d-separation-equivalence pattern corresponding to the DAG in $D$*

2. *Orient any unoriented edges which originate from, or terminate in, a node positively marked in ng of $D$, in the orientation given by the DAG in $D$*

3. *Finally, orient any edges which follow from the orientations given in the previous step and d-separation-equivalence, according to the rules derived by Meek (1995).*

*We say that a mixed graph is an **ngDAG** pattern if it represents some **ngDAG**.*

An ngDAG pattern is similar in many respects to d-separation-equivalence patterns. For example, we have the following result:

**Lemma 1** *An **ngDAG** pattern is a chain graph.*

The proof is given in the Appendix.

Our main result connects ngDAG patterns with distribution-equivalence in mixed Gaussian and non-Gaussian models in the same way that d-separation-equivalence patterns are associated with distribution-equivalence in purely Gaussian models:

**Theorem 1** *Two **ngDAGs** are distribution-equivalent if and only if they are represented by the same **ngDAG** pattern.*

The proof of this theorem is provided in the Appendix. The important point is that we now know exactly which models are indistinguishable from each other on the basis of observational data alone.

As a simple illustration, in Figure 1d we show the `ngDAG` pattern representing the `ngDAG` corresponding to the generating model of Figure 1a. Note that the `ngDAG` pattern is more informative than the d-separation-equivalence pattern of Figure 1c. Nevertheless, there are still two `ngDAG`s (leftmost two in Figure 1b) which cannot be distinguished based on non-experimental data.

Henceforth in the paper we shall use the terms *ngDAG pattern* and *distribution-equivalence pattern* interchangably.

## 5  PC-LINGAM

Although an important goal in this study was to look at the theoretical aspects of identifying DAGs in mixed Gaussian / non-Gaussian acyclic linear causal models, an equally significant objective is to give a practical method with which to infer models from a finite data set. Although there are a number of possible approaches, we here give a simple combination of independence-based techniques and the ICA-based method. The method, termed `PClingam`, consists of three steps:

1. Use methods based on conditional independence tests to estimate the d-separation-equivalence class within which the generating model lies. In particular, we advocate using the PC algorithm (Spirtes et al. 1993) which is computationally efficient even for a large number of variables. Note that, for linear models, to obtain the d-separation-equivalence class it is sufficient to identify the zero partial correlations in the data, as these depend only on the linear coefficients and the variances of the disturbances (and *not* on non-Gaussianity aspects of the distributions). However, since the data may well be signficantly non-Gaussian, non-parametric tests should optimally be used to find the zero partial correlations.

2. For each DAG $G$ in the estimated d-separation-equivalence class:

   (a) Estimate the coefficients $b_{ij}$ using ordinary least-squares regression. (Note that this provides consistent estimates regardless of non-Gaussianity of the variables.)
   
   (b) Calculate the corresponding residuals $e_i$ and rescale them to zero mean and unit variance for each $i$
   
   (c) Calculate the corresponding ICA objective function
   $$U_f = \sum_i \left(E\{f(e_i)\} - k\right)^2 \qquad (2)$$
   where $k$ is the expected value of $f$ applied to a zero-mean, unit variance Gaussian variable, i.e. $k = E\{f(g)\}$, $g \sim \mathcal{N}(0,1)$. In the ICA literature, many different choices of $f$ have been utilized; here we suggest simply taking the absolute value function $f(e_i) = |e_i|$, giving
   $$U = \sum_i \left(E\{|e_i|\} - \sqrt{2/\pi}\right)^2 \qquad (3)$$
   Of course, since we only have samples we have to take the sample mean rather than the expectation.

3. Select the highest-scoring DAG $G_{opt}$ from Step 2 and apply a statistical test for normality for each of the corresponding residuals $e_i$. Using Definition 4, compute and return the `ngDAG` pattern representing the `ngDAG` $(G_{opt}, ng)$ where $ng$ is the vector indicating those residuals whose normality was rejected by the normality tests.

The objective function $U$ is commonly used in ICA as a measure of the non-Gaussianity of a random variable, and it can be shown to give a consistent estimator for finding independent components under weak conditions (Hyvärinen et al. 2001). ICA estimation is closely related to choosing the right DAG because statistical independence of the estimated residuals is a necessary condition for the correct model: Any DAG for which the estimated residuals are not independent violates the assumptions of the model (see Section 2) and hence cannot be the data-generating DAG. On the other hand, any DAG which results in statistically independent residuals represents one valid model that could have generated the data.

Note that if we could disregard sampling effects, distribution-equivalent models would attain exactly the same value of $U$. However, in the practical case of a finite sample this is not the case, thus Step 3 in the `PClingam` algorithm is required to identify the correct distribution-equivalence class.

The method as presented above has at least a couple of shortcomings. One is that, for any given function $f(e_i)$ used, there always exist distributions which are non-Gaussian yet are not distinguished from the Gaussian by this measure. This is a well-known issue in ICA which fortunately tends to have little practical significance since few such distributions are encountered in practice. If needed, non-parametric Gaussianity measures could be used to remedy this potential problem.

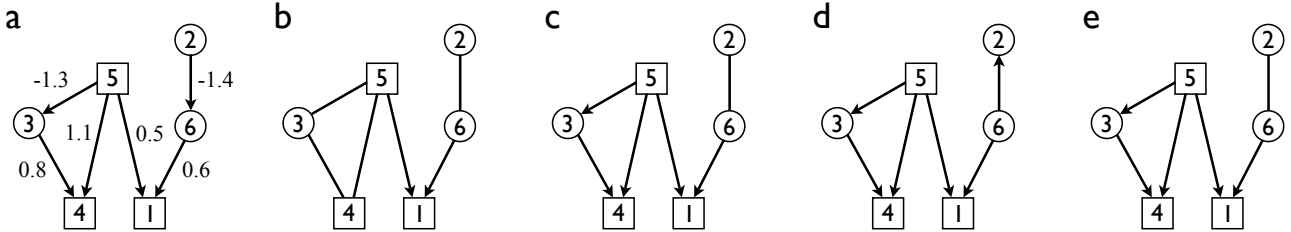

Figure 2: One of the networks used in the simulations. Variables with non-Gaussian disturbances are shown in squares, while those with Gaussian disturbances are plotted as circles. **(a)** True data-generating model. **(b)** True d-separation-equivalence pattern. **(c)** True distribution-equivalence pattern. **(d)** Estimated DAG $G_{opt}$. **(e)** Estimated distribution-equivalence pattern. See main text for details.

Naturally, in some cases many of the disturbances may be slightly non-Gaussian yet sufficiently close to Gaussian that the available samples may not be sufficient to distinguish the two and utilize the information for determining causal directions in the model. Of course, this is not a shortcoming of this particular method but is a more general phenomenon.

Another important limitation is that the ICA objective function given above will only provide a proper comparison of different DAGs for which the residuals $e_i$ are linearly uncorrelated. This is guaranteed to be the case when the search is in the correct d-separation-equivalence class, but if in Step 1 of the procedure we select a too simple model (i.e. containing too few edges) then the estimated disturbances may be linearly correlated and the objective function misleading. Thus, it might be wise to include a term penalizing linear correlations such as is used in maximum likelihood estimation of ICA (Hyvärinen et al. 2001). However, to keep our method as simple as possible, we have omitted such a penalty term in this paper.

## 6 SIMULATIONS

In this section we report on simulations used to test the performance of the `PClingam` method. First, we tested the ability of the non-Gaussianity objective function (3) of Step 2 and the normality tests of Step 3 of `PClingam` to identify the correct `ngDAG` pattern (distribution-equivalence class) when the true d-separation-equivalence pattern was known. In other words, we tested how well the algorithm would function if Step 1 of the method worked flawlessly. Subsequently, we experimented with the full method incorporating the necessary estimation of the d-separation-equivalence class (Step 1).

Figure 2a displays one of the models used to test the procedure. The disturbance distributions of variables $X_1$ and $X_5$ were a standard Gaussian the values of which were squared (but keeping the original sign) while the disturbance of $X_4$ was produced in a similar way but instead raising the values to the third power. The disturbances of $X_2, X_3,$ and $X_6$ were Gaussian. The disturbance variables were scaled such that their variances ranged from 1.0 to 3.0. A sample of 1000 data vectors was generated from the model.

Figure 2b shows the true d-separation-equivalence pattern of the model in (a). The equivalence class consists of 12 different DAGs. However, the non-Gaussianity of the disturbances of $X_1, X_4,$ and $X_5$ means that there are actually only 2 DAGs which are distribution-equivalent; these are represented by the distribution-equivalence pattern of Figure 2c. Figure 2d shows the DAG $G_{opt}$ found by Step 2 of `PClingam` from the data, when the true d-separation-equivalence class was given to the algorithm. An Anderson-Darling test for normality (Anderson and Darling 1954) gave the $p$-values 0.000, 0.3145, 0.2181, 0.000, 0.000, and 0.0197 for the corresponding residuals $e_1$ to $e_6$. Inferring a residual to be non-Gaussian when $p < 0.01$ in Step 3 of the method produced the `ngDAG` pattern of Figure 2e, which turns out identical to the true `ngDAG` pattern in (c).

This basic procedure was repeated 20 times, with the results summarized in Table 1a. In each simulation, we randomly generated a linear acyclic causal model over 6 variables, with each variable randomly chosen to have either a Gaussian or a non-Gaussian disturbance. The non-Gaussian distributions used were those mentioned above as well as a Student's t (2 degrees of freedom), a bimodal Mixture of Gaussians $(0.5\mathcal{N}(-2,1) + 0.5\mathcal{N}(2,1))$, a log-normal distribution (exponentiated standard normal) and a uniform distribution. The true d-separation-equivalence pattern was input to the algorithm, to test the functioning of the `PClingam` method when the correct pattern is selected in Step 1. The panel shows how often a specific type of true edge (in the true distribution-equivalence pattern) gave rise to a specific type of estimated edge (in the estimated distribution-equivalence pattern). Rows cor-

Table 1: Summary of the simulations employing various methods for inferring the d-separation-equivalence class in Step 1 of `PClingam`. Each table is a confusion matrix of arcs in the true distribution-equivalence patterns vs arcs in the estimated distribution-equivalence pattern. See main text for details.

**a** Using the true d-sep-equiv pattern

|   | ✗ | — | → | ← |
|---|---|---|---|---|
| ✗ | 185 | 0 | 0 | 0 |
| — | 0 | 12 | 2 | 0 |
| → | 0 | 0 | 61 | 0 |
| ← | 0 | 0 | 0 | 40 |

**b** PC

|   | ✗ | — | → | ← |
|---|---|---|---|---|
| ✗ | 183 | 0 | 0 | 2 |
| — | 0 | 12 | 2 | 0 |
| → | 9 | 1 | 43 | 8 |
| ← | 3 | 2 | 6 | 29 |

**c** CPC

|   | ✗ | — | → | ← |
|---|---|---|---|---|
| ✗ | 183 | 0 | 1 | 1 |
| — | 0 | 12 | 2 | 0 |
| → | 9 | 0 | 50 | 2 |
| ← | 3 | 1 | 2 | 34 |

**d** GES

|   | ✗ | — | → | ← |
|---|---|---|---|---|
| ✗ | 173 | 0 | 8 | 4 |
| — | 0 | 10 | 2 | 2 |
| → | 2 | 0 | 51 | 8 |
| ← | 1 | 0 | 1 | 38 |

respond to the true edges, columns to estimated ones. Optimally all off-diagonal elements would be zero. It can be seen that the results are close to perfect; the method misclassifies two undirected edges as directed, but correctly estimates all others.

These simulations confirm that the `PClingam` method works well at least when the d-separation-equivalence class can reliably be estimated. But in practice, with finite datasets, there may be significant errors in inferring the d-separation-equivalence class. The degree to which this affects the algorithm is an important practical issue.

Thus, in further simulations, we applied several different methods for learning d-separation-equivalence patterns from the simulated data, as Step 1 in the `PClingam` method. The methods we compared were the PC algorithm (Spirtes et al. 1993), the Conservative PC algorithm (Ramsey et al. 2006), and the GES algorithm (Chickering 2002). Panels b-d of Table 1 summarize the results. Although all of the methods assumed Gaussianity when learning the d-separation-equivalence pattern, the results are still quite encouraging, and a clear majority of edges were correctly estimated.

# 7  FUTURE WORK

While the theoretical aspects of identifiability are solved, at least a couple of important issues regarding the estimation of the model from finite samples remain.

First and foremost, non-parametric methods for identifying zero partial correlations in non-Gaussian settings should be used so as to obtain better estimates of the appropriate d-separation-equivalence class within which to search. Although the methods developed for Gaussian variables seem to work relatively well in our partly non-Gaussian setting, it is likely they will be outperformed by methods that take into account the possibility of non-Gaussian distributions.

Another important question is how to make the procedure scalable to data involving many (tens or even hundreds of) variables. Although the current approach relies on a brute-force enumeration of all DAGs in the d-separation-equivalence class, it would not be difficult to adapt the method to do a local search among DAGs in an equivalence class. The extent to which such a method would be hampered by local maxima is unknown.

# 8  SUMMARY

The discovery of linear acyclic causal models is a topic which has been thoroughly investigated in the last two decades. Both the Gaussian and the fully non-Gaussian special cases are well understood, but the general mixed case has not been previously discussed. In this paper we have provided a complete characterization of distribution-equivalence and a practical estimation method in this setting.

### Acknowledgements

The authors wish to thank Clark Glymour for helpful and stimulating discussions. P.O.H. was funded by a postdoctoral researcher grant from the University of Helsinki.

# APPENDIX

**Proof of Lemma 1:** Select any ngDAG $D$ represented by the ngDAG pattern $P$ in question. Now, for each variable $X_i$ in $D$ with a non-Gaussian disturbance, add two auxiliary variables both of which have no parents and both of which have $X_i$ as their only child. Now, consider what the d-separation-equivalence pattern looks like for this *augmented* graph. Any edge in $D$ *into* such an $X_i$ may be oriented in the d-separation-equivalence pattern on the basis of the resulting unshielded collider at $X_i$. Furthermore, any edge in $D$ *out of* such an $X_i$ may similarly be oriented on the basis of the lack of an unshielded collider. Thus, all edges either originating from, or terminating in, a variable with a non-Gaussian disturbance will be oriented in the d-separation-equivalence pattern of the augmented graph. Additionally, any edges whose orientations can be deduced as a result of knowing the newly oriented edges are oriented as well. Hence, adding the auxiliary variables has exactly the same effect as simply orienting any edges connected to a non-Gaussian variable.

Because d-separation-equivalence patterns are always chain graphs (Andersson et al. 1997), the d-separation-equivalence pattern for the augmented graph is a chain graph. Since each of the auxiliary variables is connected to the rest of the graph only by a single oriented edge, removing these variables cannot change the chain graph property of the graph. Hence, the distribution-equivalence pattern obtained by orienting any edges originating from, or terminating in, variables with non-Gaussian variables (and subsequently orienting any edges deduced from these) has to be a chain graph. This completes the proof.

**Proof of Theorem 1:** We will first prove that if two ngDAGs $D_1$ and $D_2$ are distribution-equivalent then they must share the same ngDAG pattern. Since distribution-equivalence implies d-separation-equivalence it is clear they have to share the d-separation-equivalence pattern. Thus we need to show that it is always possible to correctly orient any edges directly connected to any variable with a non-Gaussian disturbance.

Consider a linear acyclic causal model $M_1$ which instantiates $D_1$. Let us for simplicity of notation assume that the variables have been named $X_1, \ldots, X_n$ such that $j < i$ whenever $X_j$ is an ancestor of $X_i$. Then it is possible to collect the linear coefficients $b_{ij}$ which represent the direct causal effects into a *lower-triangular* matrix $\mathbf{B}$ such that we have $\mathbf{x} = \mathbf{Bx} + \mathbf{e}$, where $\mathbf{x}$ collects the observed variables $X_1, \ldots, X_n$ and $\mathbf{e}$ denotes the disturbance variables. Note that we have for simplicity of notation and without loss of generality assumed the constants in (1) are equal to zero. Solving for $\mathbf{x}$ we obtain $\mathbf{x} = \mathbf{Ae}$ where $\mathbf{A} = (\mathbf{I} - \mathbf{B})^{-1}$ is the *reduced-form* matrix (representing the total effects between the variables), which here is lower-triangular due to the above causal ordering of the variables. By the assumptions of the model, the components of $\mathbf{e}$ are mutually independent. This means that our generating model is an ICA model, although several (potentially even all) of the components may be Gaussian. Although it is impossible to completely estimate $\mathbf{A}$ when there are two or more Gaussian components, it is well known (Hyvärinen et al. 2001) that the basis vectors (columns of $\mathbf{A}$) corresponding to the non-Gaussian components are identifiable (except for the standard indeterminancy of permutation and scaling), as is the covariance matrix of the (generally multidimensional) Gaussian component which groups all Gaussian disturbances.

Thus each non-Gaussian disturbance in the generating model essentially gives us a vector $\mathbf{a}_i$ containing the total effects of its corresponding variable $X_i$ on the other variables. As no variable can affect its nondescendants, and as we have assumed faithfulness, the set of non-zero entries of each such vector exactly represents the union of the correct corresponding variable and all its descendants. If we could somehow know to which observed variable $X_i$ each non-Gaussian basis vector $\mathbf{a}_i$ should be paired we could easily orient all edges connecting to that variable as follows: Any observed variable $X_j$ which we know (by d-separation) to be connected to $X_i$ but which has a zero entry in $\mathbf{a}_i$ has to be a parent of $X_i$. Similarly, any observed variable $X_j$ which we know (by d-separation) to be connected to $X_i$ but which has a non-zero entry in $\mathbf{a}_i$ has to be a child of $X_i$.

But is it always possible to identify to which observed variable a given basis vector should be connected? We now show that this is indeed so. Consider what happens if we try to build a model in which a basis vector $\mathbf{a}_i$ is paired with a descendant $X_j$ of $X_i$, rather than with the correct choice $X_i$. Because we are restricted to acyclic models, the new model may be represented by a reduced-form matrix $\mathbf{A}'$ which is lower-triangular for the new variable order (but not for the old order, because in this new model $X_i$ is necessarily a descendant of $X_j$). Note also that, when the rows are identically ordered, the column $\mathbf{a}_i$ must be equal to column $\mathbf{a}'_j$ of $\mathbf{A}'$. Now, in the new model, $X_i$ has to be represented by some disturbance variable, so there must be a column $\mathbf{a}'_i$ of $\mathbf{A}'$ which has a non-zero entry in the row corresponding to $X_i$. Furthermore, $\mathbf{a}'_i$ must have zeros wherever $\mathbf{a}_i$ has zeros. These properties imply that $\mathbf{a}'_i$ cannot be expressed as a linear combination of the $\{\mathbf{a}_k\}_{k \neq i}$: Since $\mathbf{a}_i$ cannot be used, one would have to depend on $\mathbf{a}_k$ corresponding to $X_k$ which are ancestors of $X_i$ to properly represent the non-zero entry corresponding to $X_i$ in $\mathbf{a}'_i$. But among these there necessarily has to be an $\mathbf{a}_k$ which corresponds to a (relative) source $X_k$ whose effect cannot be cancelled out by the others to lead to a required zero in $\mathbf{a}'_i$. Thus $\mathbf{a}'_i$ is not in $\text{span}(\{\mathbf{a}_k\}_{k \neq i})$. This implies that $\text{span}(\{\mathbf{a}_k\}_{k \neq i}) \neq \text{span}(\{\mathbf{a}'_k\}_{k \neq j})$ and hence the covariance implied by $\mathbf{A}'$ cannot equal that implied by $\mathbf{A}$ and the two models cannot be distribution-equivalent. Hence to represent the correct distribution one has to pair basis vectors corresponding to non-Gaussian components to the correct observed variable, allowing the identification of the orientation of all edges directly connected to that observed variable.

Next we need to prove that two `ngDAG`s which share the same `ngDAG` pattern are distribution-equivalent. Call the two `ngDAG`s $D_1$ and $D_2$. Since they are represented by the same `ngDAG` pattern they may only differ in terms of the orientations of arrows within each chain component of the chain graph that is the `ngDAG` pattern. (A chain component is a maximal connected set of nodes such that there is no oriented edge between any pair of nodes.) Let $V_c$ stand for the set of variables (necessarily all with Gaussian disturbances) making up any given chain component. First, note that we can focus our attention on only the variables $V_c$ and any immediate parents of any of these: If we can match any conditional distribution $P(V_c \mid \text{pa}(V_c))$ of a model $M_1$ represented by $D_1$ using a model $M_2$ represented by $D_2$ then we can always match the full joint distribution over all variables. Next consider the parents of the chain component. It is clear that each such parent must be connected to *all* the variables of the chain component, otherwise some edges in the component could be oriented using d-separation-equivalence.

Select any parametrization $M_1$ of $D_1$, and call the resulting reduced-form matrix *between the variables $V_c$ only* $\mathbf{A}_1$. Because of the old result that d-separation-equivalence among Gaussian variables implies distribution-equivalence we know that there is a parametrization $M_2$ of $D_2$ such that the corresponding reduced-form matrix is $\mathbf{A}_2$, and we have $\mathbf{A}_1 \mathbf{A}_1^T = \mathbf{A}_2 \mathbf{A}_2^T$. Now consider the effect of a parent. Denoting the weigths of the parent onto the variables $V_c$, in model $M_1$, by the vector $\mathbf{w}_1$, we have $V_c = \mathbf{A}_1(\mathbf{e} + \mathbf{w}_1 p)$ where $\mathbf{e}$ represents the independent Gaussian disturbances of the $V_c$ and $p$ the value of the parent. Thus the conditional distribution of the $V_c$ given the value $p$ of the parent is a Gaussian with mean vector $\mathbf{A}_1 \mathbf{w}_1 p$ and covariance matrix $\mathbf{A}_1 \mathbf{A}_1^T$. For model $M_2$ we similarly have a Gaussian with mean $\mathbf{A}_2 \mathbf{w}_2 p$ and covariance matrix $\mathbf{A}_2 \mathbf{A}_2^T$. As noted above, because of d-separation-equivalence, these covariance matrices can be made identical with a suitable choice of $\mathbf{A}_2$. Then, selecting $\mathbf{w}_2 = \mathbf{A}_2^{-1} \mathbf{A}_1 \mathbf{w}_1$ yields identical means as well. (Note that the reduced-form matrices are always invertible.). Also note that additional parents always add independently, and the weights can be selected in the above fashion for each parent separately. Thus, in summary, for any chain component on which models $D_1$ and $D_2$ have conflicting edge orientations, for any parametrization $M_1$ of $D_1$ it is always possible to find a parametrization $M_2$ of $D_2$ such that the conditional distribution of the chain component given its parents is identical. Thus the two models represent the exact same set of joint distributions over all the variables. This concludes the proof of the theorem.